\documentclass[preprint,12pt]{elsarticle}

\usepackage{graphicx}%
\usepackage{amsmath,amssymb,amsfonts}%
\usepackage{amsthm}%
\usepackage{mathrsfs}%
\usepackage[title]{appendix}%
\usepackage{xcolor}%
\usepackage{textcomp}%
\usepackage{manyfoot}%
\usepackage{booktabs}%
\usepackage{algorithm}%
\usepackage{algorithmicx}%
\usepackage{algpseudocode}%
\usepackage{listings}%
\usepackage{subfigure}
\usepackage{float}
\usepackage{subcaption}
\usepackage{multirow}
\usepackage{multicol}
\usepackage{longtable}
\usepackage{graphicx}

\journal{Neural Networks}

\begin{document}

\begin{frontmatter}



\title{Visual Place Cell Encoding: A Computational Model for Spatial Representation and Cognitive Mapping}

%% \author[1,2]{\fnm{Joergen W.} \spfx{van der} \sur{Ploeg} 

\author[1]{Chance J. Hamilton}
\ead{chamilton4@usf.edu}

\author[1]{Alfredo Weitzenfeld}
\ead{aweitzenfeld@usf.edu}

\affiliation[1]{
  organization={Bellini College of Artificial Intelligence, Cybersecurity and Computer, University of South Florida},
  addressline={4202 E Fowler Ave},
  city={Tampa},
  postcode={33620},
  state={FL},
  country={USA}
}

% \affiliation[2]{
%   organization={Department of Computer Science and Engineering, University of South Florida},
%   addressline={4202 E Fowler Ave},
%   city={Tampa},
%   postcode={33620},
%   state={FL},
%   country={USA}
% }

\begin{abstract}
    This paper presents the Visual Place Cell Encoding (VPCE) model, a biologically inspired computational framework for simulating place cell–like activation using visual input. Drawing on evidence that visual landmarks play a central role in spatial encoding, the proposed VPCE model activates visual place cells by clustering high-dimensional appearance features extracted from images captured by a robot-mounted camera. Each cluster center defines a receptive field, and activation is computed based on visual similarity using a radial basis function. We evaluate whether the resulting activation patterns correlate with key properties of biological place cells, including spatial proximity, orientation alignment, and boundary differentiation. Experiments demonstrate that the VPCE can distinguish between visually similar yet spatially distinct locations and adapt to environment changes such as the insertion or removal of walls. These results suggest that structured visual input, even in the absence of motion cues or reward-driven learning, is sufficient to generate place-cell-like spatial representations and support biologically inspired cognitive mapping.
\end{abstract}

\begin{keyword}
    Visual Place Cells, Spatial Cognition, Cognitive Mapping, Robotic Navigation
\end{keyword}

\end{frontmatter}

\section{Introduction}\label{sec:intro}

Spatial representation and navigation are fundamental cognitive functions in biological systems, enabling organisms to interpret their environment and guide behavior. In mammals, the hippocampus plays a central role in this process through the activity of place cells—neurons that exhibit location-specific firing patterns as an animal navigates through space~\cite{okeefe1976place, moser2017spatial}. These place cells integrate multisensory information, including visual, vestibular, and proprioceptive cues, to form spatially selective and stable representations~\cite{jeffery2007integration, save2000contribution}.

Among the sensory modalities, visual input has been shown to play a critical role in anchoring and modulating place cell activity. Empirical studies have demonstrated that modifying visual landmarks can alter place field stability and induce remapping~\cite{chen2013vision, knierim1995place}, suggesting that structured visual information contributes significantly to allocentric spatial coding. Similar findings in non-mammalian species, such as bees navigating by landmark geometry~\cite{cartwright1983landmark}, reinforce the idea that environmental appearance alone can support spatial representation.

These biological insights have motivated the development of computational models describing how place cell–like activity can arise from visual input alone, without relying on motion signals or reward-based learning. The present work contributes to this effort by introducing a biologically inspired model in which visual place cells emerge from appearance-based clustering, providing a framework for studying how visual structure alone can support spatial tuning and cognitive mapping.

We introduce the \textit{Visual Place Cell Encoding} (VPCE), a computational model designed to simulate place cell activation using only visual inputs acquired from a mobile robot equipped with an onboard camera. The VPCE investigates whether biologically relevant spatial coding can be achieved through unsupervised organization of visual features by use of image processing clustering techniques. We define visual place cells in the VPCE as cluster centroids in a high-dimensional feature space, with activation governed by a radial basis function that reflects the similarity between current visual input and learned visual structure.

This model builds upon other existing work in biologically inspired navigation and clustering-based spatial representation~\cite{thrun2001clusterMCL, kim2003dynamicClusterMCL, mohamed2022incrementalClustering, gonzalez2015coordinatedClustering}, but is distinct in its focus: the VPCE is not a localization or mapping system, but a framework for testing whether visual appearance alone can drive place-cell-like spatial responses.We evaluate whether this model produces activation patterns that correlate with spatial localization—capturing relationships such as proximity, orientation alignment, boundary separation, and adaptability to environmental changes, all of which are characteristic of place cell activity in biological systems.

The remainder of this paper is organized as follows. Section~\ref{sec:related_work} surveys relevant biological findings on place cells and reviews computational models that inform the design of visually driven spatial representations, as well as clustering techniques in cognitive mapping and localization. Section~\ref{sec:methods} presents the architecture of the VPCE model, describing the feature extraction process, the use of clustering to define visual place cells, and the computation of activation patterns. Section~\ref{sec:experiments} outlines the experimental setup and presents results that evaluate the model’s ability to generate activation patterns that correlate with spatial localization, boundary separation, and environmental adaptation. Section~\ref{sec:discussion} presents the conclusions and discussion on results, and Section~\ref{sec:future_work} presents future work for extending the VPCE framework in neurocomputational modeling and robot navigation.

\section{Related Work}\label{sec:related_work}

Place cells in the hippocampus encode spatial locations and support goal-directed behavior by forming a cognitive map of the environment. The Visual Place Cell Encoding (VPCE) model defines a new computational framework for place cell formation, where visual inputs are processed and clustered in feature space to generate place fields. This section reviews key biological and computational models relevant to visual-based place cell formation, clustering mechanisms, and adaptive navigation, related to the VPCE model.

\subsection{Biological Place Cells}

Hippocampal place cells are neurons that become active when an animal occupies specific locations in its environment, effectively forming a cognitive map crucial for spatial navigation and memory~\cite{okeefe1976place}. These cells integrate multimodal sensory information to establish and maintain their spatial firing patterns~\cite{chen2013vision,jeffery2007integration}.

Visual cues play a particularly strong role in anchoring and stabilizing place cell activity. Chen et al.~\cite{chen2013vision} showed that visual inputs significantly contribute to the stability of place fields, supplementing path integration by providing external reference points. Early work by Knierim et al.~\cite{knierim1995place} further demonstrated that altering the stability or position of visual landmarks can cause place fields to shift or remap, reinforcing the role of vision in spatial coding.

The importance of external sensory structure in guiding navigation and memory extends beyond rodents. Cartwright and Collett~\cite{cartwright1983landmark} showed that bees rely on the angular size and bearing of landmarks for navigation, highlighting how visual cues support allocentric encoding in diverse species. Their work suggests that environment-derived visual structure can serve as the basis for spatial representations, a principle central to the VPCE model.

While vision is often dominant, place cells are influenced by other sensory inputs as well. Save et al.~\cite{save2000contribution} found that olfactory and tactile cues contribute to place field stability, especially in environments with reduced visual information. Jeffery~\cite{jeffery2007integration} argued that the integration of multiple sensory modalities supports robust and flexible spatial coding.

Moreover, interactions between place cells and other spatially tuned neurons particularly grid cells contribute to spatial precision. Moser et al.~\cite{moser2008place} proposed that grid cells in the entorhinal cortex provide a metric scaffold that informs hippocampal place fields. In humans, Rolls~\cite{rolls2023hippocampal} extended this idea by identifying spatial view cells that respond to specific visual perspectives, suggesting the hippocampus may encode spatial context more richly than physical position alone.

The VPCE model aims to capture key aspects of these biological findings. By clustering high-dimensional visual features, it tests whether place cell-like activation patterns can arise solely from visual structure without reference to self-motion or multisensory integration. Through this design, it provides a computational framework for evaluating proximity encoding, structural differentiation, and environmental adaptation in visually driven systems.

\subsection{Computational Place Cell Models}

The VPCE model introduces a computational framework in which spatially selective activity emerges from unsupervised clustering of high-dimensional visual features. Each place cell is defined by a cluster centroid in feature space, and its activation is computed using a radial basis function modulated by the intra-cluster distance. This design allows the VPCE to generate graded spatial responses based solely on visual input, forming a representational map without requiring learned transitions, temporal sequences, or external supervision.

Several computational models of place cells have focused on predictive coding frameworks, including the successor representation (SR), which encodes expected future state occupancy~\cite{stachenfeld2014sr}. These models typically rely on transition statistics learned through repeated exploration to generate place-like activity patterns. In contrast, the VPCE organizes spatial representations based on perceptual structure, where place cell tuning emerges from the geometry of clustered visual features rather than predictions over state transitions.

Other models have explored hierarchical reinforcement learning as a mechanism for multi-scale spatial abstraction. For example, Chalmers et al.~\cite{chalmers2024hierarchical} and Scleidorovich et al.~\cite{scleidorovich2022multi} showed that hierarchical planning structures can lead to spatial codes at varying resolutions. In VPCE, variation in receptive field size emerges naturally from clustering dynamics—specifically, differences in intra-cluster variance result in a spatial code that reflects both fine-grained and coarse environmental structure.

Biologically grounded models have also investigated how plasticity mechanisms, such as Hebbian learning and recurrent feedback, shape the formation and stability of place fields over time~\cite{hasselmo2020models,yuan2015entorhinal}. Rather than relying on such learning mechanisms, the VPCE provides a controlled framework for evaluating whether place-cell-like activation patterns can emerge from clustered visual representations alone.

In contrast to models that simulate neural competition, such as self-organizing maps~\cite{amari1977dynamics}, the VPCE achieves spatial differentiation by partitioning the visual feature space using $k$-means clustering. This approach enables the emergence of structured receptive fields while maintaining computational simplicity and interpretability.

Some computational models have focused on how place cells adapt to context and changes in environmental structure. The VPCE captures this dimension through its use of intra-cluster distances, which modulate place cell sensitivity and lead to heterogeneous receptive field sizes. This diversity resembles functional differences proposed along the dorsal-ventral axis of the hippocampus~\cite{contreras2018ventral}, though VPCE does not impose anatomical structure explicitly.

The VPCE takes a visual-centric approach to place field generation, leveraging appearance-based structure rather than motion-derived signals such as path integration or grid cell metrics~\cite{gaussier2007path}. This design avoids drift effects associated with cumulative path integration errors~\cite{milford2010ratslam}, while emphasizing the utility of visual features in establishing spatial codes. Rather than encoding directional transitions between states, as in models with transition cells~\cite{cuperlier2007transition}, the VPCE evaluates whether spatial relationships and environmental boundaries can be inferred directly from visual similarity.

Although abstracted from biophysical and synaptic mechanisms, the VPCE aligns with broader neuroscientific principles suggesting that structured sensory input can give rise to spatial codes. For instance, Polykretis and Michmizos~\cite{polykretis2022astrocytes} highlighted astrocytic influences on synaptic dynamics, while Moser et al.~\cite{moser2017spatial} described the progression from spatially localized activity to contextually modulated place fields in the hippocampus. VPCE echoes these ideas by generating spatial codes through perceptual similarity rather than geometric coordinates or multisensory integration.

This modeling approach also resonates with early behavioral studies showing that visual structure alone can guide navigation. Cartwright and Collett~\cite{cartwright1983landmark} demonstrated that bees navigate based on the angular size and bearing of landmarks, while Zeno et al.~\cite{zeno2016review} reviewed the role of structured sensory feedback in neurobiologically inspired navigation. By clustering appearance-based visual input to form spatially informative units, the VPCE contributes a biologically motivated computational perspective on spatial encoding in artificial and natural systems.

\begin{figure}[t!]
    \centering
    \includegraphics[width=\textwidth]{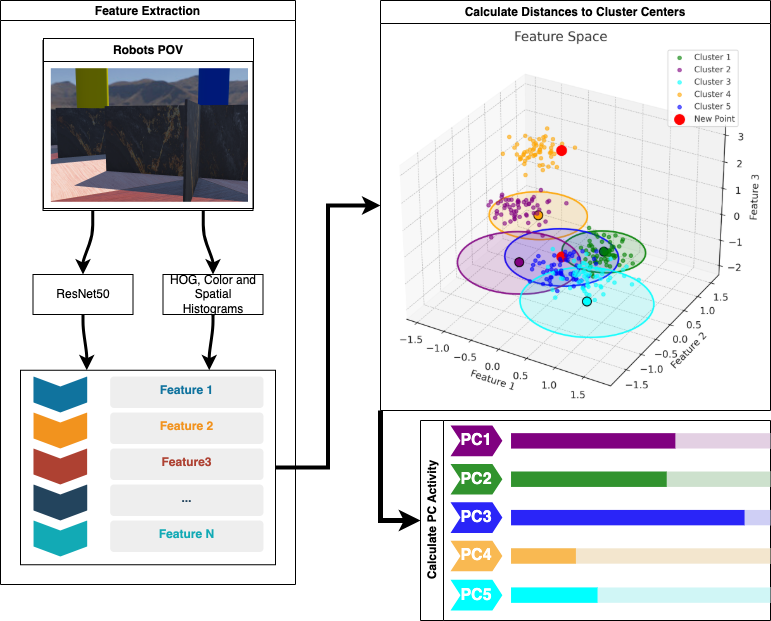}
    \caption{Illustration of the Visual Place Cell Encoding (VPCE) model. Feature vectors derived from visual observations are embedded in a high-dimensional feature space and clustered. Each cluster centroid represents a place field center. When a new observation is processed, the VPCE computes distances to all centroids, and activation levels are determined using a radial basis function. Feature vectors that are close in the feature space produce higher activations, while dissimilar observations result in weaker responses.}
    \label{fig:vpce_overview}
\end{figure}

\section{Methods}\label{sec:methods}

This section outlines the complete procedure for constructing and evaluating the Visual Place Cell Encoding (VPCE) model. Section~\ref{sec:formationVPC} introduces the overall concept of visual place cells and summarizes the modeling approach. Section~\ref{sec:vpce_modeling} details the activation framework, including radial basis function computation and normalization. Finally, Section~\ref{sec:VPCE_pipline} integrates these components into a unified pipeline, illustrating how visual inputs are transformed into spatially structured place cell representations.

\subsection{Formation of Visual Place Cells}\label{sec:formationVPC}

The Visual Place Cell Encoding (VPCE) model transforms visual observations into structured spatial representations by clustering high-dimensional feature vectors. As illustrated in Figure~\ref{fig:vpce_overview}, the process begins with feature extraction from images captured during agent exploration. These features are projected into a shared feature space, where unsupervised clustering identifies regions of visual similarity. Each resulting cluster defines a place cell, with its activation field centered at the cluster centroid, and with its spatial sensitivity determined by the spread of points within the cluster. When a new observation is encountered, its activation pattern is computed based on its similarity to the precomputed place cell centers using a radial basis function. We show that these place cell activations are strongly correlated with physical location within the environment. 

This section details the core components of the VPCE model. Section~\ref{sec:FeatureExtraction} describes the feature extraction process, which combines deep and handcrafted descriptors to embed visual observations. Section~\ref{sec:clustering_process} outlines the clustering methodology used to define visual structure in the feature space. Section~\ref{sec:PlaceCells} explains how place cell centers are established and how the model computes graded activation patterns for novel inputs.

\subsubsection{Feature Extraction}\label{sec:FeatureExtraction}

Feature extraction transforms raw visual data acquired during robot exploration into structured feature vectors suitable for clustering and visual place cell formation. Two distinct feature extraction methods were employed and compared: a convolutional neural network (CNN)-based approach and a multimodal approach using hand-crafted descriptors.

The CNN-based approach utilized a pre-trained ResNet50 convolutional neural network \cite{he2016deep} with the classification layers removed, resulting in high-dimensional semantic feature vectors. CNN-based features have been widely used in image representation and clustering tasks due to their ability to capture abstract visual representations from large-scale datasets \cite{babenko2014neural}. Formally, the CNN-based feature vector is defined as:
\begin{equation}
\mathbf{f}_{\text{CNN}} = [\text{ResNet50 Features}]
\label{eq:cnn_features}
\end{equation}

The multimodal feature extraction approach combined several established hand-crafted visual descriptors. Specifically, the multimodal feature vector consisted of the following:
\begin{itemize}
    \item \textbf{Histogram of Oriented Gradients (HOG)}: Captures edge orientations and gradient structures \cite{dalal2005histograms}.
    \item \textbf{Color Histogram}: Represents color distribution information within images, commonly used to discriminate between visually distinct landmarks \cite{swain1991color}.
    \item \textbf{Spatial Histogram}: Encodes the spatial distribution of intensity values, preserving structural information within visual scenes \cite{lazebnik2006beyond}.
\end{itemize}

The multimodal feature vector is formally represented as:
\begin{equation}
\mathbf{f}_{\text{Multimodal}} = [\text{HOG},\, \text{Color Histogram},\, \text{Spatial Histogram}]
\label{eq:multimodal_features}
\end{equation}

These extracted feature vectors were subsequently utilized in clustering analyses to identify visual place cell centers. An overview of the visual place cell formation pipeline is presented in Figure~\ref{fig:vpce_overview}, which illustrates the steps from feature extraction to cluster-based place cell activation.

\subsubsection{Clustering Process}\label{sec:clustering_process}

To identify representative points within the extracted feature space, a clustering algorithm was employed. While various clustering methods exist such as DBSCAN, hierarchical clustering, and spectral clustering the K-Means clustering algorithm \cite{macqueen1967some} was selected due to its computational simplicity, scalability, and explicit generation of centroid-based clusters \cite{jain2010data}. These centroids are directly utilized as receptive field centers for the Visual Place Cell Encoding (VPCE), providing spatially meaningful representations derived from the feature vectors.

K-Means clustering partitions feature vectors into \( k \) clusters by iteratively minimizing the sum of squared Euclidean distances between feature vectors and their respective cluster centroids. Given a dataset \( X = \{x_1, x_2, ..., x_n\} \), K-Means clustering aims to minimize the objective function:
\begin{equation}
J = \sum_{i=1}^{k} \sum_{x_j \in C_i} \| x_j - \mu_i \|^2
\end{equation}
where \( C_i \) represents the set of feature vectors belonging to cluster \( i \), and \( \mu_i \) denotes the centroid of cluster \( i \).

Determining the optimal number of clusters \( k \) and selecting the most suitable feature representation were addressed through a quantitative experimental evaluation, which is described in detail within the results section.

\subsubsection{Establishment of Place Cell Centers}\label{sec:PlaceCells}

The centroids resulting from the K-Means clustering process serve as representative points in the extracted feature space. These centroids correspond to key locations within the visual representation, effectively defining specific receptive fields for the Visual Place Cell Encoding (VPCE). By establishing this set of representative feature points, each centroid can subsequently be related to a distinct spatial region in the robot's environment, thereby bridging the transition from feature extraction and clustering to visual place cell modeling.

\subsection{Visual Place Cell Modeling}\label{sec:vpce_modeling}

Visual place cells are computational units whose activations encode spatial information derived from the robot's visual inputs. After representative centroids are established via K-Means clustering in the extracted feature space, each centroid defines the receptive field center of a visual place cell. These centroids correspond to visually meaningful regions in the environment, forming the basis for spatial encoding. The VPCE model defines a population of such place cells, where each unit’s activity reflects the visual similarity between a new observation and the learned spatial representations. This section details the mechanism used to compute activation values and how these values form a coherent spatial code. Section~\ref{sec:RBF} describes the use of radial basis functions for modeling activation as a function of distance to cluster centers. Section~\ref{sec:PCActivation} introduces a normalization step to ensure interpretability across the ensemble. Section~\ref{sec:Spatial} explains how the resulting activation patterns are used to represent spatial relationships and analyze structural differentiation in the environment.

\subsubsection{Radial Basis Function Activation}\label{sec:RBF}

To computationally model the activation of visual place cells, we employ radial basis functions (RBFs), which have been shown to accurately model place cell activation in the hippocampus \cite{burgess1994model}. The strength of a visual place cell’s activation is determined by the Euclidean distance between an input feature vector \(\mathbf{f}\) and the centroid \(\mathbf{c}_i\), scaled according to the spatial extent or spread of the cluster associated with that centroid.

The scaling parameter \(\alpha_i\) for each visual place cell is defined as the maximum intra-cluster distance:

\begin{equation}
\alpha_i = \max_{\mathbf{f} \in \text{Cluster } i} \| \mathbf{f} - \mathbf{c}_i \|
\label{eq:scaling_parameter}
\end{equation}

The activation of the \(i^{th}\) visual place cell is defined as:

\begin{equation}
A_i(\mathbf{f}) = \exp\left( -\frac{\| \mathbf{f} - \mathbf{c}_i \|^2}{2 \alpha_i^2} \right)
\label{eq:place_cell_activation}
\end{equation}

This RBF-based formulation ensures a smoothly decreasing response as the input feature vector moves away from the centroid, providing a localized and continuous spatial encoding.

\subsubsection{Activation Normalization}\label{sec:PCActivation}

To ensure consistency and balance among visual place cell activations, a min-max normalization is applied across all activations:

\begin{equation}
\tilde{A}_i = \frac{A_i - \min(A)}{\max(A) - \min(A)}
\label{eq:min_max_normalization}
\end{equation}

This normalization prevents any single place cell from disproportionately dominating the ensemble response. The normalized activations collectively represent how strongly the input feature vector corresponds to the various spatial regions defined by the visual place cells.

\subsubsection{Spatial Representation}\label{sec:Spatial}

The final activation pattern across the ensemble reflects the input feature’s alignment with the underlying spatial structure of the environment. High activation values indicate that the input feature vector closely corresponds to the spatial region represented by the visual place cell, providing a distributed encoding that supports navigation and localization tasks.

\subsection{Visual Place Cell Encoding Pipeline}\label{sec:VPCE_pipline}

\begin{figure}[t!]
    \centering
    \subfigure[]{\includegraphics[width=0.85\textwidth]{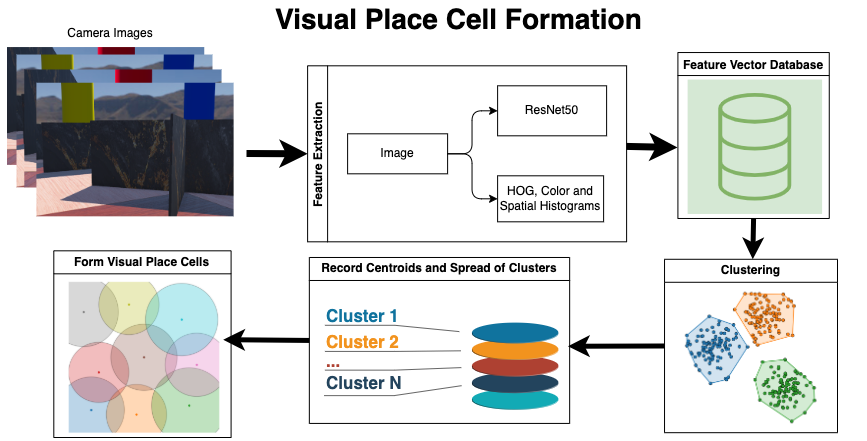}\label{fig:vpce_pipeline_a}}\\[1ex]
    \subfigure[]{\includegraphics[width=0.85\textwidth]{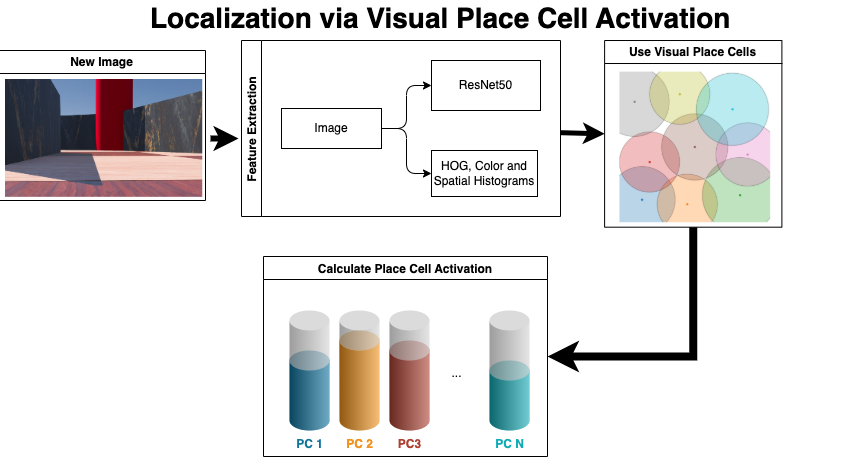}\label{fig:vpce_pipeline_b}}
    \caption{Overview of the Visual Place Cell Encoding (VPCE) model. \textbf{(a)} Construction of the visual place cell ensemble: POV images collected from a navigating agent are processed through a feature extraction pipeline combining ResNet50 and low-level descriptors (HOG, color histograms, spatial histograms). Extracted feature vectors are clustered to define centroids and intra-cluster spreads that specify visual place fields. \textbf{(b)} Place cell activation during deployment: A new image is processed through the same pipeline and compared to learned centroids. Place cell activations are computed using radial basis functions centered at each centroid and scaled by intra-cluster spread, producing a graded activation pattern based on visual similarity.}
    \label{fig:vpce_pipeline}
\end{figure}

Figure~\ref{fig:vpce_pipeline} illustrates the full Visual Place Cell Encoding (VPCE) pipeline, comprising two main stages: ensemble formation (Figure~\ref{fig:vpce_pipeline}\textbf{(a)}) and activation during deployment (Figure~\ref{fig:vpce_pipeline}\textbf{(b)}).

In the first stage, a dataset of point-of-view (POV) images is collected as the agent explores its environment. Each image is passed through a feature extraction module that combines deep embeddings from a pre-trained ResNet50 with handcrafted descriptors, including histogram of oriented gradients (HOG), color histograms, and spatial histograms. The resulting feature vectors are clustered in the visual feature space using $k$-means. The centroids and intra-cluster spreads define the ensemble of visual place cells and their receptive fields.

In the second stage, a new image is processed through the same feature extraction pipeline and compared to the learned centroids. Activation of each place cell is computed using a radial basis function centered at each centroid and scaled by its intra-cluster spread, resulting in a graded activation pattern that reflects the similarity between the current input and the stored visual place fields.

This pipeline allows the VPCE to generate structured, location-sensitive activation patterns without relying on odometry, path integration, or metric coordinates. The resulting representation supports remapping, boundary encoding, and serves as a biologically inspired foundation for downstream tasks in navigation and cognitive mapping.

\section{Experiments and Results}\label{sec:experiments}

This section presents a series of experiments designed to evaluate the representational properties of the Visual Place Cell Encoding (VPCE) model. Section~\ref{sec:data_collection} outlines the data generation process used for testing, including the simulation settings and visual input collection. Section~\ref{sec:ClusterQual} evaluates the quality of clustering in the visual feature space using internal metrics. Section~\ref{sec:SpatialEncoding} examines whether the resulting place cell activations preserve spatial proximity relationships. Section~\ref{sec:obstacles} investigates the model’s ability to differentiate locations across physical barriers. Finally, Section~\ref{sec:EnvirChanges} assesses how well the VPCE representation generalizes when structural changes are introduced into the environment.

\subsection{Data Collection and Preprocessing}\label{sec:data_collection}
To evaluate the spatial encoding capabilities of the VPCE model, a dataset of visual observations was collected from a controlled simulation environment. These observations serve as the input for feature extraction and subsequent clustering. This section outlines the process used to generate and structure the data. Section~\ref{sec:SimEnv} describes the simulation environment used to emulate realistic agent perception and spatial structure. Section~\ref{sec:DataAcqu} details the random exploration policy used to acquire diverse point-of-view images across the environment. The resulting dataset provides the foundation for generating place cell representations from visual input alone.

\subsubsection{Simulation Environment}\label{sec:SimEnv}

\begin{figure}[htbp]
    \centering
    \subfigure[Open environment without obstacles.]{
        \includegraphics[width=0.47\textwidth]{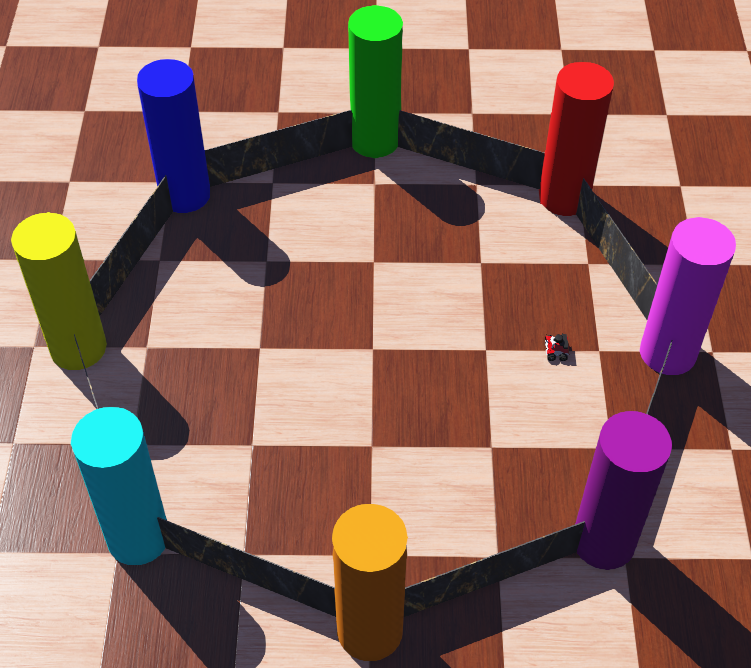}
        \label{fig:open-environment}
    }
    \hfill
    \subfigure[Environment containing internal walls as obstacles.]{
        \includegraphics[width=0.45\textwidth]{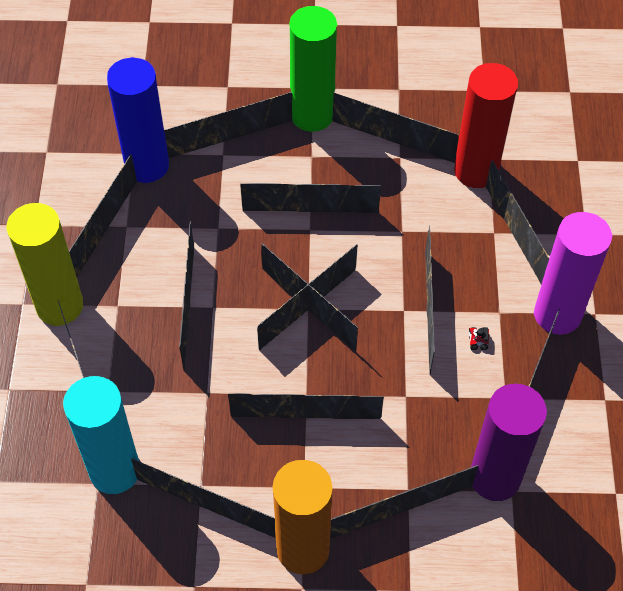}
        \label{fig:walled-environment}
    }
    \caption{Simulation environments utilized in this study, each containing eight unique landmarks (colored cylinders). The Husarion ROSbot equipped with LiDAR and RGB-D camera sensors is shown within each environment.}
    \label{fig:simulation-environments}
\end{figure}

Data collection was conducted within the Cyberbotics Webots simulator using Husarion's ROSbot platform, equipped with a LiDAR sensor, an RGB-D camera, and additional onboard sensors. Two distinct simulation environments were designed: an open environment devoid of obstacles, featuring eight uniquely colored cylindrical landmarks, and a second environment containing internal walls as obstacles, again featuring eight identical landmarks. Figure~\ref{fig:simulation-environments} illustrates both simulation environments along with the ROSbot as simulated in Webots.

\subsubsection{Data Acquisition via Random Exploration}\label{sec:DataAcqu}

To gather visual data necessary for forming the Visual Place Cell Encoding (VPCE), the robot performed a randomized exploration of the environments. The exploration strategy consisted of eight predefined actions, each corresponding to a translation of $0.3\,m$ along one of eight equally spaced headings (increments of $\frac{\pi}{4}$ radians), with $0$ radians aligned along the positive x-axis of the environment. The robot executed $1000$ random actions, selecting each action from a slightly modified uniform probability distribution biased towards the previous action to encourage coherent directional exploration.

The detailed random exploration procedure is summarized in Algorithm~\ref{alg:random-exploration}.

\begin{algorithm}[htbp]
    \caption{Random Exploration for Data Acquisition}
    \label{alg:random-exploration}
    \begin{algorithmic}[1]
        \State Initialize robot at random position and orientation.
        \State Define action set $A = \{0, \frac{\pi}{4}, \frac{\pi}{2}, \frac{3\pi}{4}, \pi, \frac{5\pi}{4}, \frac{3\pi}{2}, \frac{7\pi}{4}\}$.
        \State Select initial action $a_{\text{prev}}$ uniformly from $A$.
        \For{$i=1$ to $1000$}
            \State Select action $a_i$ from modified uniform distribution, biased towards $a_{\text{prev}}$.
            \State Use onboard LiDAR to verify feasibility of action $a_i$.
            \If{action $a_i$ is feasible (collision-free)}
                \State Rotate robot to heading $a_i$.
                \State Move forward $0.3\,m$.
                \For{each heading $a_j \in A$}
                    \State Rotate robot to heading $a_j$.
                    \State Capture and store RGB-D image.
                    \State Record positional data $(x, y, \theta)$.
                \EndFor
                \State Set $a_{\text{prev}} = a_i$.
            \Else
                \State Select a new action in the next iteration.
            \EndIf
        \EndFor
        \State Shuffle the collected images.
        \State Split data: $80\%$ for place cell formation, $20\%$ for evaluation.
    \end{algorithmic}
\end{algorithm}

\subsection{Evaluation of Clustering Quality}\label{sec:ClusterQual}
\subsubsection{Objective}

The purpose of this experiment is to evaluate the impact of feature type and number of clusters on the quality of clusters formed using K-Means clustering. Clustering quality directly influences the effectiveness of the Visual Place Cell Encoding (VPCE) in encoding spatial information, as high-quality clusters provide distinct and stable receptive fields in the feature space.

While increasing the number of clusters typically improves clustering performance according to established clustering metrics \cite{jain2010data}, the primary goal of this experiment is to determine how the choice of feature type CNN-based or multimodal affects clustering quality. Identifying the feature type that produces higher-quality clusters will help establish the most effective representation for place cell formation.

\subsubsection{Methodology}

Clustering was performed using the 80\% training set of the dataset. The following configurations were tested:

\begin{itemize}
    \item CNN-based features with cluster counts of $k = 10, 20, 100, 500$.
    \item Multimodal features with cluster counts of $k = 10, 20, 100, 500$.
\end{itemize}

K-Means clustering was used to identify representative feature points, as described in Section~\ref{sec:clustering_process}. The cluster centroids produced by this process serve as the receptive field centers for the visual place cells.

\subsubsection{Evaluation Metrics}

The quality of clustering was evaluated using three standard performance metrics commonly applied in clustering analysis.

\begin{itemize}
    \item The \textbf{Silhouette Score} metric measures how similar a point is to its own cluster compared to neighboring clusters. Higher silhouette scores indicate better-defined and more compact clusters~\cite{rousseeuw1987silhouettes}.
    
    \item The \textbf{Davies-Bouldin Index (DBI)} metric quantifies the ratio of intra-cluster distances to inter-cluster separation. Lower values of DBI suggest more compact and well-separated clusters~\cite{davies1979cluster}.
    
    \item The \textbf{Calinski-Harabasz Index (CHI)} metric evaluates the ratio of between-cluster variance to within-cluster variance. Higher values of CHI reflect better-defined and more cohesive clusters~\cite{calinski1974dendrite}.
\end{itemize}

These metrics provide a comprehensive assessment of clustering quality by capturing intra-cluster compactness, inter-cluster separation, and overall cluster definition.

\subsubsection{Results}

Table~\ref{tab:clustering_metrics} presents the clustering performance results for multimodal and CNN-based features. The results reflect how clustering quality changes with the number of clusters.

For the multimodal feature extraction approach, the Silhouette Score increases as the number of clusters increases, reaching a peak value of 0.185 at 500 clusters. The Davies-Bouldin Index (DBI) decreases with increasing cluster count, indicating more compact clusters, while the Calinski-Harabasz Index (CHI) decreases as expected due to the increased number of clusters.

For the CNN-based feature extraction approach, the Silhouette Score and CHI values are lower than those observed with multimodal features, suggesting poorer clustering quality. However, the DBI decreases with increasing clusters, indicating improved separation and cohesion at higher cluster counts.

These results suggest that multimodal features consistently produce higher-quality clusters compared to CNN-based features. The higher Silhouette Score and lower DBI for multimodal features, particularly at higher cluster counts, support the hypothesis that multimodal features provide a more informative and structured representation of the simulated environment we used for experimentation.

\begin{table}[h!]
\centering
\begin{tabular}{c c c c c c c}
\toprule
\textbf{\# Clusters} & \multicolumn{2}{c}{\textbf{Silhouette}} & \multicolumn{2}{c}{\textbf{DBI}} & \multicolumn{2}{c}{\textbf{CHI}} \\
\cmidrule(lr){2-3} \cmidrule(lr){4-5} \cmidrule(lr){6-7}
& \textbf{MM} & \textbf{CNN} & \textbf{MM} & \textbf{CNN} & \textbf{MM} & \textbf{CNN} \\
\midrule
10  & 0.130 & 0.018 & 2.172 & 4.808 & 581.15 & 89.21 \\
20  & 0.107 & 0.010 & 2.494 & 4.470 & 347.89 & 55.30 \\
100 & 0.095 & 0.026 & 2.340 & 3.529 & 108.21 & 19.27 \\
500 & 0.185 & 0.093 & 1.564 & 2.216 & 44.34  & 2.22  \\
\bottomrule
\end{tabular}
\caption{Comparison of clustering performance metrics for multimodal (MM) and CNN-based (CNN) feature extraction approaches.}
\label{tab:clustering_metrics}
\end{table}

\subsection{Assessment of Spatial Encoding Capacity}\label{sec:SpatialEncoding}
\subsubsection{Objective}

The goal of this experiment is to evaluate the ability of the Visual Place Cell Encoding (VPCE) to encode spatial information. Specifically, the objective is to establish whether feature vectors derived from spatially close locations and similar orientations produce similar VPCE activation patterns, while spatially distant vectors or those with different orientations produce distinct activation patterns. Additionally, the experiment aims to assess how the choice of feature type (CNN-based or multimodal) and the number of place cells ($k = 10, 20, 100, 500$) influence the quality of spatial encoding.

\subsubsection{Methodology}

A total of 200 reference feature vectors were randomly selected from the dataset. For each reference vector, four sets of five additional feature vectors were constructed based on spatial proximity and orientation. The first set, referred to as \textbf{Close Same}, included vectors that were spatially close and aligned with the same orientation. The second set, \textbf{Far Same}, included vectors that were spatially distant but aligned with the same orientation. The third set, \textbf{Close Different}, included vectors that were spatially close but misaligned in orientation. The final set, \textbf{Far Different}, included vectors that were both spatially distant and misaligned in orientation.

VPCE activation patterns were generated using the place cell centers established from clustering in the previous experiment. The experiment was conducted using cluster counts of $k = 10, 20, 100, 500$ for both CNN-based and multimodal feature vectors. For each of the four groupings, the average similarity of the corresponding VPCE activation patterns was computed across all reference samples. This allowed a direct comparison of the spatial encoding capacity of different feature types and cluster configurations.

\subsubsection{Evaluation Metrics}\label{sec:spatial_encoding_metrics}

The similarity of VPCE activation patterns was evaluated using the following measures:

Cosine similarity measures how aligned two activation vectors are, where values closer to 1 indicate higher similarity \cite{singhal2001modern}. Pearson correlation coefficient measures the linear relationship between two activation vectors, with higher values indicating stronger correlation \cite{benesty2009pearson}. Euclidean distance measures the straight-line distance between two activation vectors, where smaller values indicate greater similarity.

\begin{equation}
\text{Cosine Similarity} = \frac{\mathbf{A}_i \cdot \mathbf{A}_j}{\| \mathbf{A}_i \| \| \mathbf{A}_j \|}
\end{equation}

\begin{equation}
\text{Pearson Coefficient} = \frac{\sum_{i=1}^n (A_i - \bar{A})(B_i - \bar{B})}{\sqrt{\sum_{i=1}^n (A_i - \bar{A})^2 \sum_{i=1}^n (B_i - \bar{B})^2}}
\end{equation}

\begin{equation}
\text{Euclidean Distance} = \sqrt{\sum_{i=1}^n (A_i - B_i)^2}
\end{equation}

\begin{figure}[h!]
    \centering
    \includegraphics[width=0.9\textwidth]{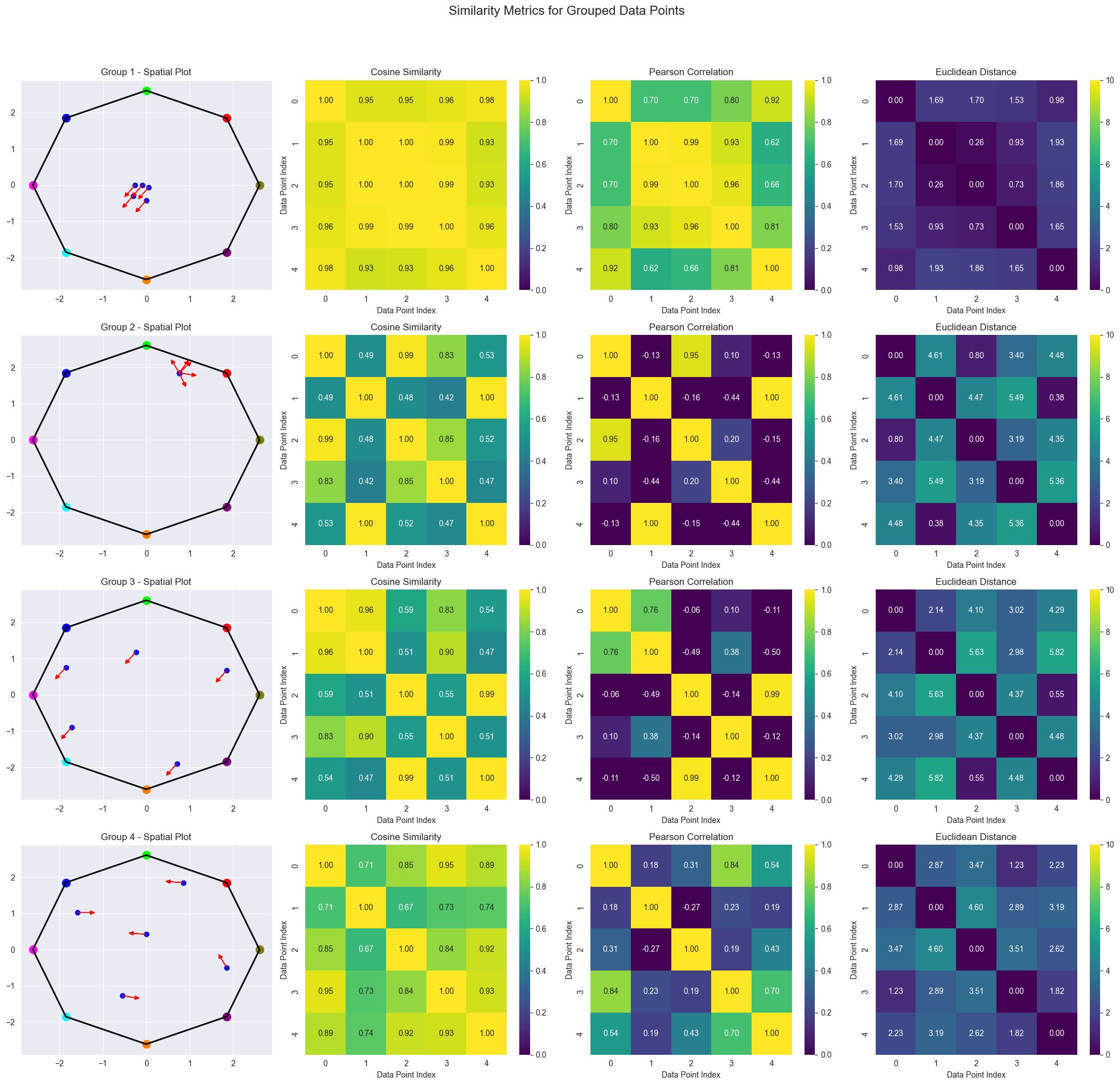}
    \caption{Similarity metrics for VPCE activation patterns across four spatial groupings. Each row represents a different group of five data points selected based on spatial arrangement and orientation. The leftmost column shows the spatial positions and headings of the data points within a bounded environment. The three similarity matrices (cosine similarity, Pearson correlation, and Euclidean distance) show pairwise comparisons of the VPCE activation patterns. Groups composed of spatially proximal and similarly oriented data points (e.g., Group 1) exhibit high cosine similarity and low Euclidean distances between activation patterns, while groups with greater spatial separation or divergent orientations (e.g., Groups 2–4) show reduced similarity and increased distance, highlighting the spatial sensitivity of the VPCE representation.}
    \label{fig:vpce_group}
\end{figure}

Figure~\ref{fig:vpce_group} illustrates the grouping configurations and corresponding similarity metrics. Each row in the figure represents a different grouping type (e.g., close same, far same), and the spatial plots show the relative positions of the grouped data points. The similarity matrices demonstrate the expected patterns: higher cosine similarity and Pearson correlation, and lower Euclidean distance for "close same" groupings; and lower similarity and higher distance for "far different" groupings. This confirms that the VPCE encodes spatial proximity and orientation consistency.

\subsubsection{Results}

Table~\ref{tab:place_cell_count_feature_type} presents the similarity of VPCE activation patterns for different spatial groupings and feature types across varying place cell counts. 

For both CNN-based and multimodal feature extraction approaches, similarity increased with the number of place cells. For CNN-based features, cosine similarity and Pearson correlation were consistently higher for Close-Same groupings, reaching near-perfect alignment at 100 place cells. Multimodal features showed similar trends, but with greater variability at higher place cell counts.

At 10 and 20 place cells, CNN-based features produced higher similarity values for Close-Same groupings compared to other configurations, with an average cosine similarity of 0.9426 and 0.9669, respectively. The difference between Close-Same and Distant-Different groupings was larger for CNN-based features, highlighting their ability to encode spatial proximity.

Multimodal features demonstrated more consistent similarity patterns across groupings but showed increased Euclidean distance at higher place cell counts. At 100 place cells, multimodal features showed reduced similarity across groupings, suggesting that increasing place cell resolution may reduce encoding efficiency for complex multimodal input.

These results suggest that CNN-based features are more effective at encoding spatial proximity and orientation at moderate place cell counts, while multimodal features may introduce greater variability at higher clustering resolutions.

\begin{table}[h!]
\centering
\resizebox{\textwidth}{!}{%
\begin{tabular}{c c c c c c c c}
\toprule
\textbf{\# Place Cells} & \textbf{Grouping} & \multicolumn{2}{c}{\textbf{Cosine Similarity}} & \multicolumn{2}{c}{\textbf{Pearson Correlation}} & \multicolumn{2}{c}{\textbf{Euclidean Distance}} \\
\cmidrule(lr){3-4} \cmidrule(lr){5-6} \cmidrule(lr){7-8}
& & \textbf{CNN} & \textbf{MM} & \textbf{CNN} & \textbf{MM} & \textbf{CNN} & \textbf{MM} \\
\midrule
\multirow{4}{*}{10} 
& Close-Same         & \textbf{0.9426} & \textbf{0.9074} & \textbf{0.8236} & \textbf{0.6845} & \textbf{0.5522 }& \textbf{0.6888} \\
& Close-Different    & 0.7902 & 0.8013 & 0.3616 & 0.3199 & 1.0326 & 1.0382 \\
& Distant-Same       & 0.7692 & 0.7583 & 0.2875 & 0.2072 & 1.1225 & 1.1782 \\
& Distant-Different  & 0.7736 & 0.7712 & 0.3719 & 0.1874 & 1.0598 & 1.2145 \\
\midrule
\multirow{4}{*}{20} 
& Close-Same         & \textbf{0.9669} & \textbf{0.9612} & \textbf{0.8844} & \textbf{0.7519} & \textbf{0.5973} & \textbf{0.6921} \\
& Close-Different    & 0.9027 & 0.8810 & 0.6299 & 0.4091 & 1.0616 & 1.2210 \\
& Distant-Same       & 0.8682 & 0.8694 & 0.5004 & 0.3215 & 1.2588 & 1.3818 \\
& Distant-Different  & 0.8706 & 0.8810 & 0.5221 & 0.3733 & 1.2670 & 1.3055 \\
\midrule
\multirow{4}{*}{100} 
& Close-Same         & \textbf{0.9950} & \textbf{0.9414} & \textbf{0.9466} & \textbf{0.7297} & \textbf{0.6671} & \textbf{1.4507} \\
& Close-Different    & 0.9862 & 0.8369 & 0.8493 & 0.3686 & 1.1330 & 2.4124 \\
& Distant-Same       & 0.9817 & 0.7857 & 0.7963 & 0.2309 & 1.3322 & 2.8421 \\
& Distant-Different  & 0.9827 & 0.8030 & 0.8060 & 0.2689 & 1.3191 & 2.7283 \\
\bottomrule
\end{tabular}%
}
\caption{Effect of place cell count on activation similarity for different feature types and spatial groupings.}
\label{tab:place_cell_count_feature_type}
\end{table}

\subsection{Spatial Differentiation in the Presence of Obstacles}\label{sec:obstacles}

\subsubsection{Objective}

The goal of this experiment is to evaluate whether the Visual Place Cell Encoding (VPCE) produces distinct activation patterns for feature vectors obtained from spatially separated regions on opposite sides of a wall. If the VPCE is encoding meaningful spatial information, activation patterns for feature vectors acquired from the same side of a wall should be more similar, while activation patterns from opposite sides should be more distinct. This experiment will establish whether the VPCE can represent spatial separation induced by physical barriers.

\subsubsection{Methodology}

A new VPCE was generated for an environment containing internal walls using the methods described in Section~\ref{sec:vpce_modeling}. Multimodal feature vectors were used to construct the model, with $k = 100$ place cells defining the ensemble.

\begin{figure}[h!]
    \centering
    \includegraphics[width=0.7\textwidth]{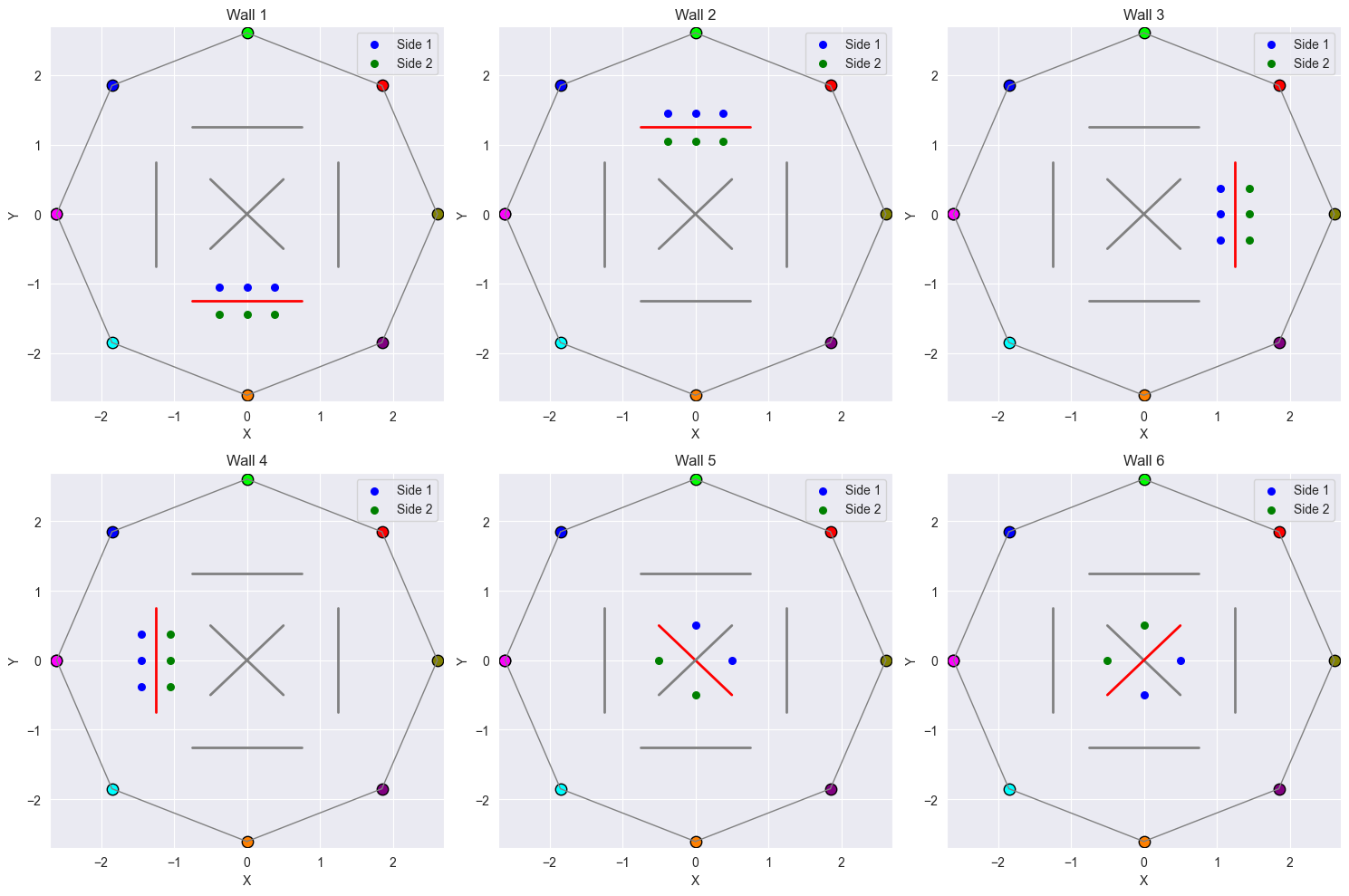}
    \caption{Example of testing points used for the spatial differentiation experiment. Colored points indicate sampling positions on each side of the wall. Red lines represent the physical barriers.}
    \label{fig:wall_positions}
\end{figure}

Feature vectors were collected from the positions shown in Figure~\ref{fig:wall_positions}. At each position, the robot captured images at eight equally spaced orientations (increments of $\frac{\pi}{4}$ radians). VPCE activation patterns were generated for each feature vector.

To evaluate the effect of spatial separation by walls, similarity was computed between feature vectors sampled from:
\begin{itemize}
    \item The same side of the wall, with the same orientation.
    \item Opposite sides of the wall, with the same orientation.
\end{itemize}

Figure~\ref{fig:wall_positions} shows the experimental setup, with sampling points on each side of the wall. Feature vectors were generated at the indicated positions and orientations, and similarity was computed between pairs of activation patterns.

\subsubsection{Evaluation Metrics}

The similarity of VPCE activation patterns was evaluated using the same metrics from Section~\ref{sec:spatial_encoding_metrics} (Cosine Similarity, Pearson Correlation, and Euclidean Distance). For each wall configuration, the average similarity between feature vectors on the same side of the wall was compared to the average similarity between feature vectors on opposite sides. Higher similarity values for same-side comparisons and lower similarity for opposite-side comparisons would indicate that the VPCE successfully encodes spatial separation induced by physical barriers.

\begin{figure}[h!]
    \centering
    \includegraphics[width=0.9\textwidth]{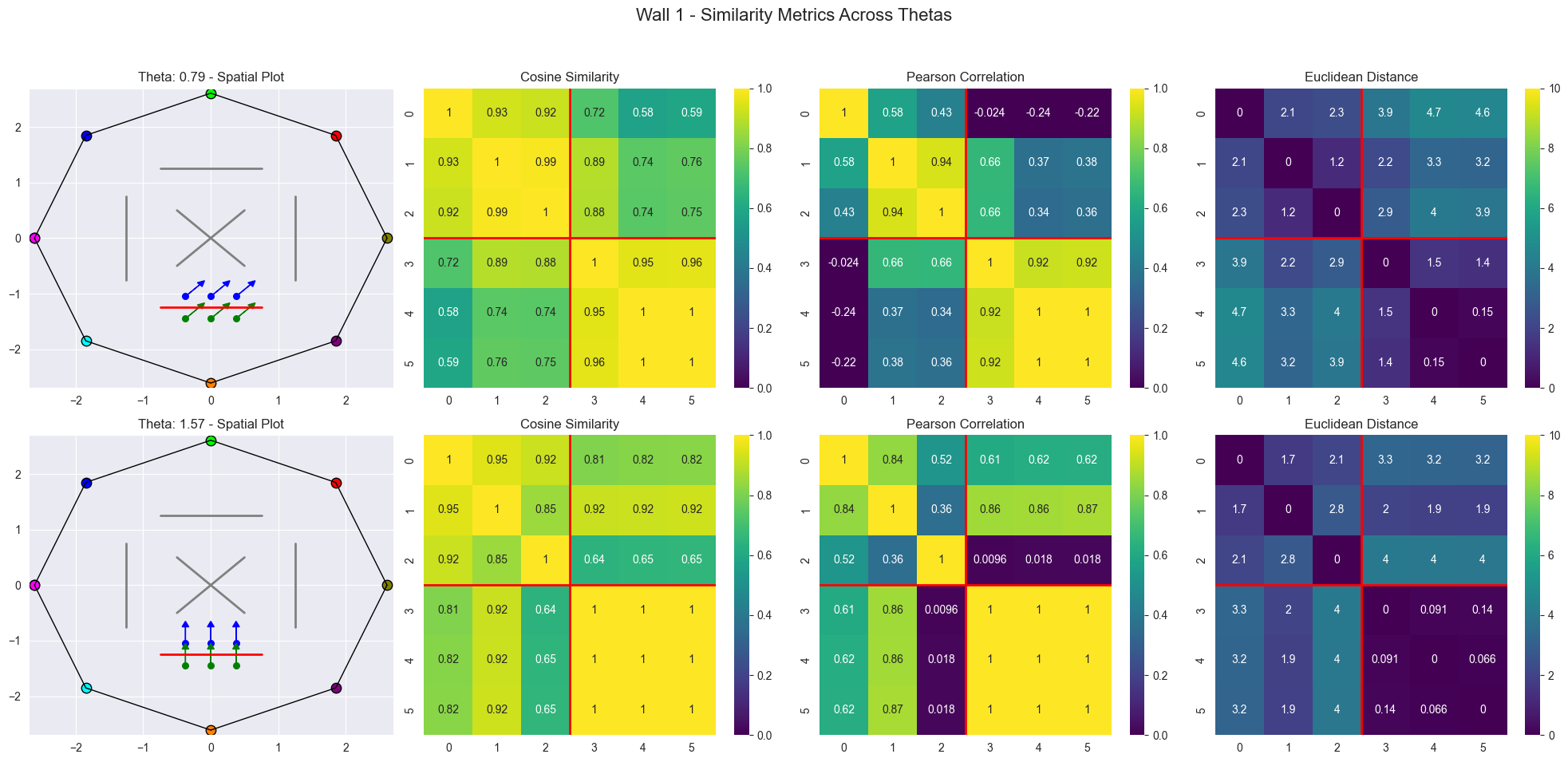}
    \caption{Similarity metrics for VPCE activation patterns across data points sampled from either the same side or opposite sides of a wall, at two fixed orientations. Each row shows a different $\theta$ value (top: $\theta = 0.79$, bottom: $\theta = 1.57$). The spatial plots show the positions and headings of the six data points—three on each side of the wall. In the similarity matrices (cosine similarity, Pearson correlation, and Euclidean distance), matrix quadrants correspond to types of comparisons: the top-left and bottom-right quadrants show same-side comparisons, while the off-diagonal quadrants compare across the wall. Higher cosine similarity and lower Euclidean distance are observed for within-side comparisons, while across-wall comparisons exhibit reduced similarity and greater distance, indicating that the VPCE encodes spatial separation even under similar visual orientations.}
    \label{fig:wall_results}
\end{figure}

Figure~\ref{fig:wall_results} shows the similarity matrices for each metric. Higher cosine similarity and lower Euclidean distance between feature vectors on the same side of the wall (upper left and lower right of matrices) indicate consistent spatial encoding. Conversely, lower cosine similarity and higher Euclidean distance for opposite side (upper right and lower left of matrices)comparisons suggest that the VPCE can distinguish spatial locations separated by walls.

\subsubsection{Results}

\begin{figure}[h!]
    \centering
    \includegraphics[width=0.6\textwidth]{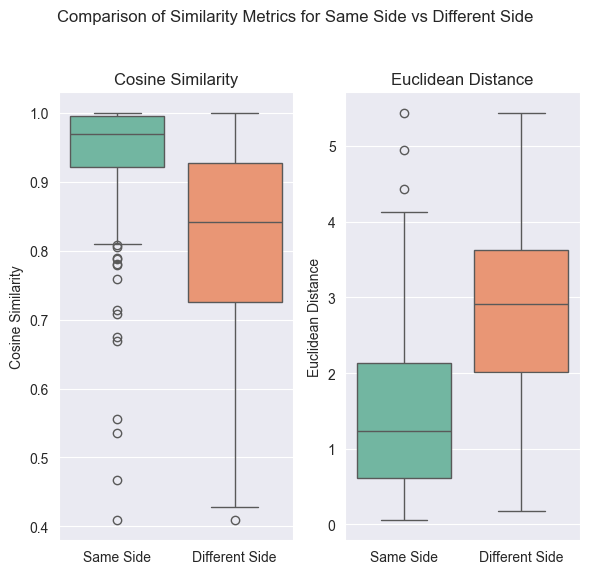}
    \caption{Comparison of similarity metrics for same-side and different-side patterns. Cosine similarity is higher and Euclidean distance is lower for same-side patterns, confirming that the VPCE encodes spatial separation imposed by physical barriers.}
    \label{fig:wall_boxplot}
\end{figure}

Figure~\ref{fig:wall_boxplot} shows the distribution of cosine similarity and Euclidean distance for VPCE activation patterns on the same and opposite sides of a wall. The goal of this experiment was to determine whether the VPCE produces distinct activation patterns for spatially separated feature vectors across a physical barrier.

Cosine similarity values were higher for feature vectors on the same side of the wall compared to different sides, while Euclidean distance was lower for same-side patterns. This suggests that the VPCE encodes spatial proximity and separation consistently with the environment’s structure.

To evaluate statistical significance, an independent two-sample t-test was performed. The results showed a statistically significant difference in cosine similarity ($t = 5.60$, $p = 1.99 \times 10^{-6}$) and Euclidean distance ($t = -6.92$, $p = 3.16 \times 10^{-8}$) between same-side and different-side patterns. The low p-values confirm that the differences are unlikely to have occurred by chance, indicating that the VPCE accurately differentiates spatial separation imposed by physical barriers.

\subsection{Adaptation to Environmental Changes}\label{sec:EnvirChanges}

\subsubsection{Objective}

The purpose of this experiment is to evaluate the ability of the Visual Place Cell Encoding (VPCE) to adapt to structural changes in the environment. Specifically, we aim to test two distinct types of environmental modifications:

\begin{enumerate}
    \item \textbf{Wall Addition:} A VPCE is formed from an environment with no internal walls. After introducing a wall into the environment, we assess whether the VPCE can develop distinct activation patterns on opposite sides of the new wall, consistent with the spatial differentiation observed in the previous experiment without the need of reforming VPCE.
    
    \item \textbf{Wall Removal:} A VPCE is formed from an environment with internal walls. After removing a wall, we evaluate whether the VPCE maintains similar activation patterns for closely grouped place cells despite the absence of the wall without the need of reforming VPCE.
\end{enumerate}

If the VPCE is encoding meaningful spatial information, the addition of a wall should lead to more distinct activation patterns on opposite sides, while the removal of a wall should preserve similarity between previously grouped place cells.

\subsubsection{Methodology}

Two existing VPCE models were considered for this experiment:

\begin{enumerate}
    \item \textbf{Open Environment VPCE:} A multimodal VPCE with $k = 100$ place cells formed using data collected from an open environment with no internal walls.
    \item \textbf{Internal Wall VPCE:} A multimodal VPCE with $k = 100$ place cells formed using data collected from an environment containing internal walls.
\end{enumerate}

For the wall addition case, the Open Environment VPCE was used to generate activation patterns before and after introducing a wall into the environment. For the wall removal case, the Internal Wall VPCE was used to generate activation patterns before and after removing a wall. 

Feature vectors were collected from the same spatial locations before and after the environmental change, with images captured at eight equally spaced orientations (increments of $\frac{\pi}{4}$ radians) at each location. VPCE activation patterns were computed for the pre-change and post-change feature vectors to evaluate how well the VPCE adapted to the structural modifications.

\subsubsection{Evaluation Metrics}

The similarity of VPCE activation patterns was evaluated using the same metrics from Section~\ref{sec:spatial_encoding_metrics} (Cosine Similarity, Pearson Correlation, and Euclidean Distance). 

For the wall addition case, higher similarity for feature vectors on the same side of the new wall and lower similarity for opposite sides would indicate successful adaptation to the structural change. Conversely, for the wall removal case, consistent similarity between previously grouped feature vectors despite the removal of the wall would suggest that the VPCE retains meaningful spatial encoding even when the environment becomes less constrained.

\begin{figure}[h!]
    \centering
    \includegraphics[width=0.9\textwidth]{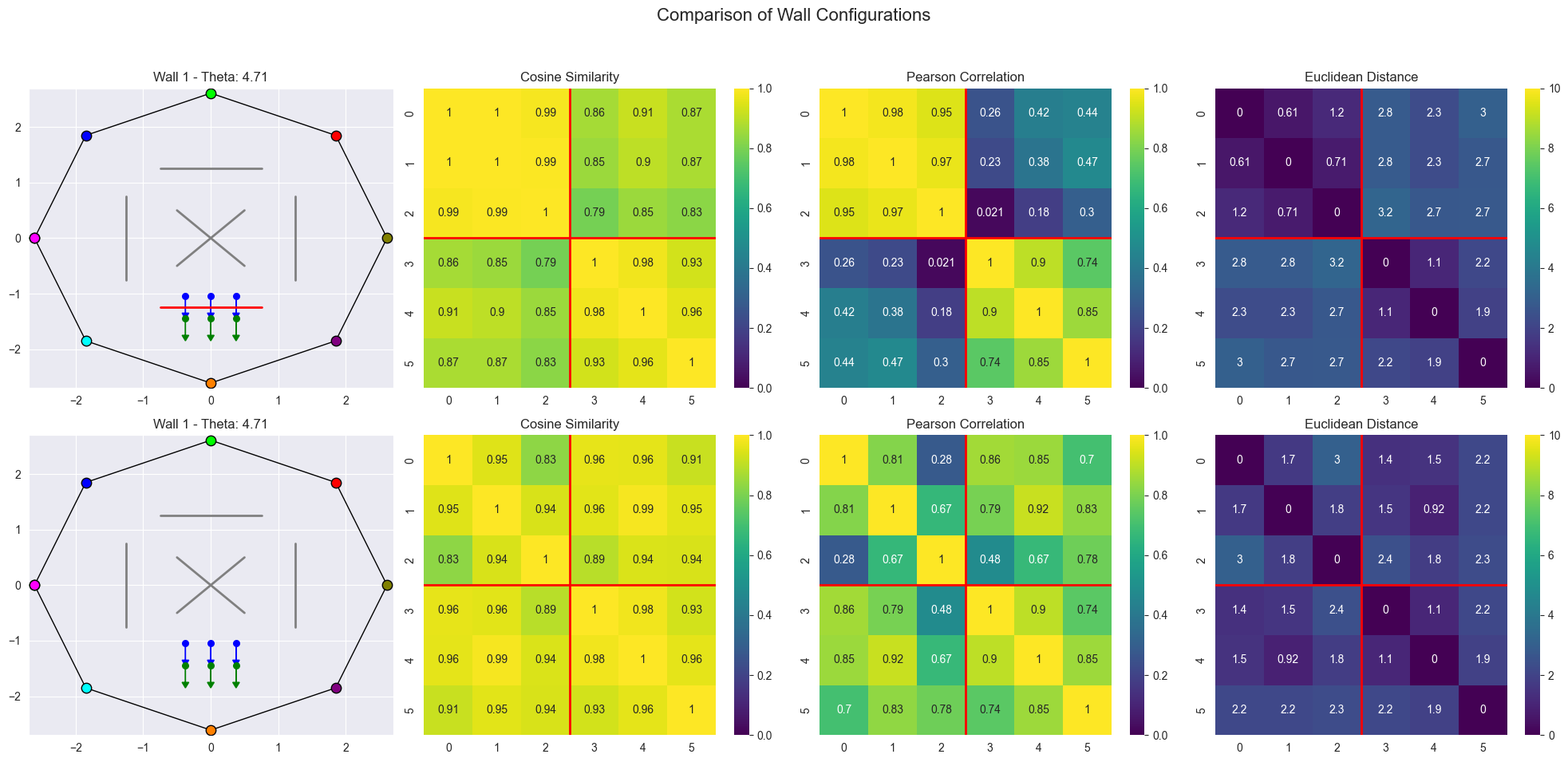}
    \caption{Comparison of VPCE activation pattern similarity before and after the removal of Wall 1. Each row shows results at a fixed orientation ($\theta = 4.71$), with the top row corresponding to the configuration before wall removal and the bottom row showing the configuration after removal. The spatial plots depict the location and heading of six data points—three from each side of the original wall. The similarity matrices (cosine similarity, Pearson correlation, and Euclidean distance) compare pairwise VPCE activation patterns. Matrix quadrants correspond to types of comparisons: top-left and bottom-right show within-side similarity, while off-diagonal quadrants compare across the wall. Following wall removal, across-group similarity increases and Euclidean distance decreases, indicating that the VPCE representation adapts to changes in environmental structure.}
    \label{fig:wall_diff}
\end{figure}

Figure~\ref{fig:wall_diff} illustrates the effect of wall removal on VPCE activation patterns. The top row shows similarity matrices computed from VPCE activation patterns before the wall was removed, while the bottom row shows similarity matrices after the wall was removed. Higher cosine similarity and Pearson correlation, and lower Euclidean distance for patterns on the same side are observed in the pre-change state. After wall removal, similarity increases across previously separated patterns, indicating that the VPCE adjusts to the new spatial configuration.

\subsubsection{Results}

To evaluate the VPCE’s ability to adapt to environmental changes, we tested two scenarios: the addition of a wall and the removal of a wall. The goal was to determine whether the VPCE could differentiate between activation patterns before and after structural changes to the environment.

\begin{figure}[t!]
    \centering
    \includegraphics[width=0.5\textwidth]{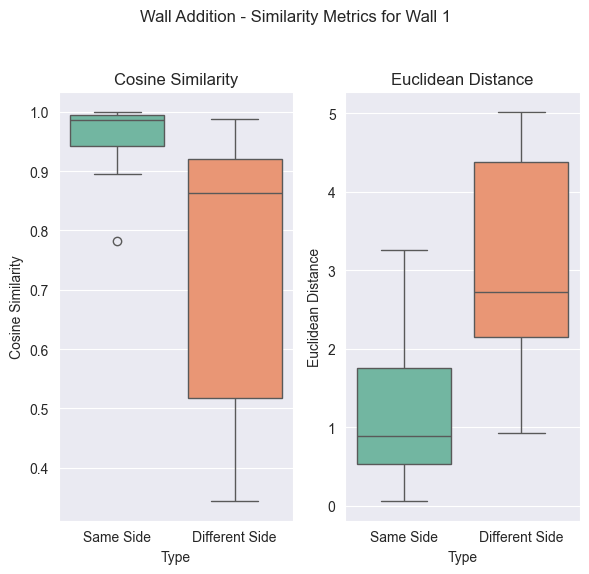}
    \caption{Comparison of similarity metrics for VPCE activation patterns following the addition of a wall. Higher cosine similarity and lower Euclidean distance for same-side patterns indicate successful spatial encoding of the barrier.}
    \label{fig:wall_add_bp}
\end{figure}

Figure~\ref{fig:wall_add_bp} shows the similarity metrics for VPCE activation patterns following the addition of a wall. Cosine similarity was significantly higher for feature vectors on the same side of the wall compared to opposite sides, while Euclidean distance was lower for same-side patterns. An independent two-sample t-test confirmed a statistically significant difference between same-side and different-side patterns for cosine similarity ($t = 3.97$, $p = 0.0003$) and Euclidean distance ($t = -5.20$, $p = 7.07 \times 10^{-6}$).

\begin{figure}[t!]
    \centering
    \includegraphics[width=0.5\textwidth]{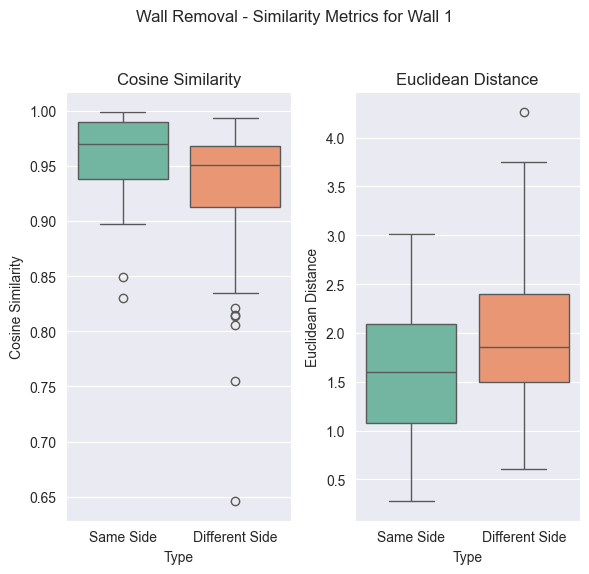}
    \caption{Comparison of similarity metrics for VPCE activation patterns following wall removal. The lack of significant difference suggests preserved spatial encoding of previously grouped patterns.}
    \label{fig:wall_remove_bp}
\end{figure}

Figure~\ref{fig:wall_remove_bp} shows the similarity metrics after wall removal. The VPCE maintained similar activation patterns for feature vectors previously separated by the wall. Cosine similarity was not significantly different ($t = 1.87$, $p = 0.070$), and Euclidean distance was also not significantly different ($t = -1.60$, $p = 0.119$). This suggests that the VPCE preserves the spatial relationship of previously grouped patterns even after structural changes.

\section{Conclusions and Discussion}\label{sec:discussion}

The Visual Place Cell Encoding (VPCE) model was developed to test whether place cell–like activity can emerge from visual input alone without access to motion information, temporal sequences, or task-driven learning. The core aim of this work is to evaluate whether spatially structured activation patterns can be generated from robot-acquired visual data using a biologically inspired mechanism. This section discusses the results of our experimental evaluation in the context of spatial coding and cognitive mapping.

\subsection{Clustering Quality and Feature Representations}

Our first experiment examined how different feature types and ensemble sizes influence clustering quality, which in turn shapes the spatial structure of the VPCE. Clustering quality improved as the number of clusters increased, though gains plateaued beyond 100 place cells. CNN-based features produced more compact and well-separated clusters, as indicated by higher silhouette scores and lower Davies-Bouldin Index values, compared to the multimodal representation. 

These results are important not as an optimization of clustering performance per se, but because cluster quality influences the spatial granularity of the resulting visual place cells. High-quality clusters create sharper and more selective receptive fields in feature space, a property desirable in simulating spatially tuned neuronal responses. The findings confirm that CNN-derived features provide a more structured basis for generating place cell representations.

\subsection{Spatial Encoding Consistency}

The second experiment evaluated whether VPCE activation patterns correlate with spatial proximity and orientation alignment. We found that observations collected from nearby positions and similar headings produced significantly more similar activation patterns than those from distant or misaligned locations. These effects were consistent across metrics (cosine similarity, Pearson correlation, and Euclidean distance), and were more pronounced in models built using CNN-based features.

This supports the hypothesis that the VPCE captures structured spatial information using appearance-based cues alone. In biological systems, place cells are known to exhibit spatially graded firing patterns modulated by both position and heading. The VPCE reproduces this behavior in silico through clustering in a visual feature space, offering a computational analog to proximity and orientation-dependent encoding in the hippocampus.

\subsection{Boundary Differentiation}

Our third experiment tested whether the VPCE could distinguish between locations on opposite sides of a physical barrier. We observed that activation patterns were more similar for samples taken on the same side of a wall and more distinct across the barrier. These differences were statistically significant, indicating that the model encoded structural separation even in the absence of changes in visual orientation.

This result parallels the phenomenon of boundary-modulated place field remapping observed in hippocampal neurons, where spatial context and environmental geometry influence firing fields. In the VPCE, such differentiation arises from clustering structure alone, demonstrating that environmental discontinuities can be reflected in purely appearance-driven spatial codes. The ability to represent boundaries is critical for cognitive map formation and spatial segmentation.

\subsection{Adaptation to Structural Changes}

The fourth experiment assessed how the VPCE responds to dynamic changes in the environment. When a wall was introduced into an initially open environment, similarity between activation patterns on either side of the new barrier decreased, reflecting a shift in spatial representation. Conversely, when a wall was removed, VPCE activation patterns remained similar to those observed before removal, preserving previous spatial associations.

This form of plasticity, achieved without retraining or re-clustering, suggests that the VPCE is capable of adapting its representations in response to structural changes while maintaining stability where appropriate. Such behavior is consistent with adaptive place cell dynamics observed in biological systems, where spatial representations can update to reflect new constraints while preserving prior structure when possible.

\subsection{Contributions}

This work introduces the Visual Place Cell Encoding (VPCE) model as a biologically inspired framework for generating spatially structured activation patterns from visual input alone. The VPCE departs from models that rely on motion cues, temporal dynamics, or learned transitions by demonstrating that visual appearance, as captured by a robot-mounted camera, can generate place cell–like responses.

Our primary contribution lies in showing that visual place cells can be modeled using appearance-based clustering in feature space, producing activation patterns that reflect key properties of biological place cells—namely, spatial proximity, orientation alignment, boundary differentiation, and adaptation to structural changes. These effects emerge without any online learning or retraining, underscoring the utility of visual structure as a basis for spatial representation.

The computational tools used CNN-based features and k-means clustering were contrasted to analyze their impact on the modeling framework proposed in this work, used in the formulation and evaluation of the place cell biologically grounded model that explains how allocentric spatial coding may arise from visual input in both artificial and natural agents.

\begin{figure}[t!]
    \centering
    \includegraphics[width=\textwidth]{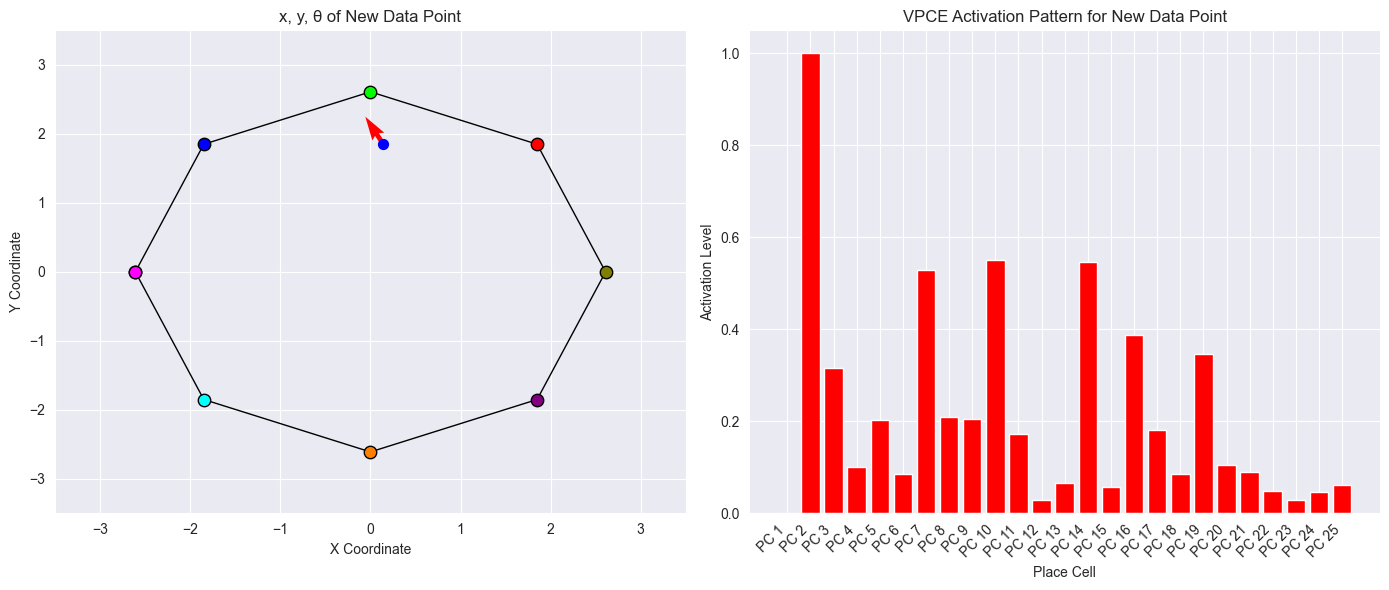}
    \caption{VPCE activation pattern for a single visual observation. Left: spatial location and heading of the robot at the time of observation. Right: activation levels for each place cell in the ensemble. While only a few cells show high activation, the majority exhibit non-zero responses due to the overlap of receptive fields in feature space and the smooth radial basis function used to compute similarity. This continuous activation reflects structural properties of the model rather than biological sparsity, and provides motivation for future thresholding mechanisms.}
    \label{fig:vpce_activation_example}
\end{figure}

While the VPCE model captures key spatial properties observed in biological place cells, one notable distinction lies in how activation is represented across the ensemble. In biological systems, place cells are typically considered active only when their firing rate surpasses a physiological threshold, resulting in sparse population responses in which only a limited number of cells are active at a given location. In contrast, the VPCE computes a non-zero activation value for every place cell using a radial basis function centered at each cluster centroid in the feature space. This results in continuous, graded activation levels across the entire ensemble even for observations that are visually dissimilar from most centroids. As illustrated in Figure~\ref{fig:vpce_activation_example}, this effect arises from the substantial overlap between receptive fields in the feature space, which leads to small but non-zero responses in distant units due to the smooth decay of the radial basis function. While this continuous activation profile enhances stability and interpretability in downstream applications, it departs from the sparse activation patterns typically observed in hippocampal recordings.

\section{Future Work}\label{sec:future_work}

The Visual Place Cell Encoding (VPCE) model presented in this work provides a foundation for investigating spatial representation through biologically inspired computation. Several key directions will extend this framework to address open questions in systems neuroscience, reinforcement learning, and robotics.

\subsection{Modeling Multi-Field Place Cell Ensembles}

Empirical studies have demonstrated that individual hippocampal place cells can exhibit multiple firing fields across an environment, particularly in large or topologically complex spaces~\cite{Fenton2008Multifield, Park2011EnsemblePlasticity}. These multi-field place cells present a challenge to traditional models that assume a unique spatial tuning per cell. The VPCE framework currently models place cells as single-field units associated with distinct clusters in visual feature space.

Future work will expand the VPCE to model multi-field ensembles by enabling place cells to associate with multiple non-contiguous regions in feature space. We plan to consider thresholding and other sparsity-promoting mechanisms to suppress weak responses and more closely match the selective firing behavior found in biological place cell ensembles. This will allow us to investigate how overlapping or context-dependent fields arise and how they contribute to ensemble-level spatial coding. A key goal is to determine whether such overlapping representations improve spatial generalization or introduce ambiguity during localization. These studies will contribute to understanding the representational flexibility of hippocampal networks and may provide a computational explanation for observed remapping phenomena.

Another key direction for extending the biological realism of the VPCE model involves investigating mechanisms for enforcing population sparsity. In contrast to the sparse firing of place cells observed in hippocampal recordings, the VPCE currently computes non-zero activation values for all cells in the ensemble. This behavior arises from the overlapping structure of receptive fields in the feature space and the smooth falloff of the radial basis function used to compute activation. To better reflect biological patterns of activity, future work will explore thresholding methods to suppress weak activations, or competitive mechanisms such as $k$-winners-take-all that restrict the number of active units. These modifications will support comparisons between continuous and sparse representations, offering insights into how activation structure affects generalization, stability, and downstream learning performance.

\subsection{Integrating VPCE with Reinforcement Learning Agents}

A central aim of biologically inspired spatial representation is to support goal-directed behavior. To that end, future work will integrate the VPCE into the state representation of reinforcement learning (RL) agents navigating obstacle-rich environments. By encoding observations through visual place cell activations, the VPCE can serve as a structured, low-dimensional input that retains spatial relationships and contextual differentiation.

This line of work will explore several dimensions of model refinement. First, we will compare alternative feature encoding strategies, including deep neural embeddings, biologically plausible visual descriptors, and hybrid methods. Second, we will investigate clustering architectures beyond standard k-means, such as hierarchical~\cite{Murtagh2012HierarchicalClustering} or graph-based clustering approaches that incorporate motion continuity~\cite{Stachenfeld2014DesignPrinciples}. These methods may better reflect the topological structure of large environments and allow place field representations to emerge at multiple spatial scales.

A critical component of this integration involves quantifying the relationship between environment size and the number of place cells required for sufficient state coverage. This scaling analysis will inform both biological interpretation and practical implementation. In addition, future extensions will incorporate biologically inspired mechanisms such as replay and preplay~\cite{Johnson2005Replay, Pfeiffer2013HippocampalPreplay}, enabling the model to simulate offline memory consolidation and predictive planning observed in hippocampal circuits.

\subsection{Deployment in Physical Robotic Systems}

To evaluate the applicability of VPCE in real-world navigation tasks, future work will involve deploying the model on a physical mobile robot platform. This will enable testing the robustness of visual place field encoding under real-world perceptual noise, changes in lighting and occlusion, and imperfect actuation. The VPCE will be coupled with onboard sensing and reinforcement learning modules to enable end-to-end learning of navigation policies in novel environments.

Key goals include assessing how well the VPCE generalizes across physical spaces and determining the level of sensory fidelity required for consistent activation patterns. The integration of biologically inspired replay and place field adaptation mechanisms will also be evaluated in the context of lifelong learning and map plasticity. This direction will bridge the gap between computational modeling and embodied cognition, demonstrating how principles of hippocampal function can inform the design of flexible, spatially aware robotic agents.

\section{Acknowledgements}

This work was funded in part by NSF IIS Robust intelligence research grant \#1703225 at the University of South Florida entitled "Experimental and Robotics Investigations of Multiscale Spatial Memory Consolidation in Complex Environments".


% \bibliography{sn-bibliography} 

% \end{thebibliography}

\bibliographystyle{elsarticle-num-names} 
\bibliography{sn-bibliography} 

\end{document}